\documentclass{article} 
\usepackage{iclr2025_conference,times}

\usepackage{amsmath,amsfonts,bm}

\def\eqref#1{equation~\ref{#1}}

\def\1{\bm{1}}

\DeclareMathAlphabet{\mathsfit}{\encodingdefault}{\sfdefault}{m}{sl}
\SetMathAlphabet{\mathsfit}{bold}{\encodingdefault}{\sfdefault}{bx}{n}

\usepackage{natbib}  
\usepackage{hyperref}
\usepackage{url}
\usepackage{graphicx}
\usepackage{booktabs}
\usepackage{longtable}

\title{Probing Perceptual Constancy in Large Vision-Language Models}

\iclrfinalcopy
\author{
  \textbf{Haoran Sun}\textsuperscript{1}\thanks{Correspondence to \texttt{haoransun.ms@gmail.com}} \quad
  \textbf{Bingyang Wang}\textsuperscript{2} \quad
  \textbf{Suyang Yu}\textsuperscript{3} \quad
  \textbf{Yijiang Li}\textsuperscript{4} \quad
  \textbf{Qingying Gao}\textsuperscript{1} \\
  \textbf{Haiyun Lyu}\textsuperscript{5} \quad
  \textbf{Lianyu Huang}\textsuperscript{6} \quad
  \textbf{Zelong Hong}\textsuperscript{7} \quad
  \textbf{Jiahui Ge}\textsuperscript{6} \quad
  \textbf{Qianli Ma}\textsuperscript{8} \\
  \textbf{Hang He}\textsuperscript{9} \quad
  \textbf{Yifan Zhou}\textsuperscript{8} \quad
  \textbf{Lingzi Guo}\textsuperscript{10} \quad
  \textbf{Lantao Mei}\textsuperscript{10} \\
  \textbf{Maijunxian Wang}\textsuperscript{11} \quad
  \textbf{Dezhi Luo}\textsuperscript{12} \quad
  \textbf{Hokin Deng}\textsuperscript{13} \\
  \textsuperscript{1}Johns Hopkins University \quad
  \textsuperscript{2}Emory University \quad
  \textsuperscript{3}University of Washington \\
  \textsuperscript{4}University of California San Diego \quad
  \textsuperscript{5}University of North Carolina at Chapel Hill \\
  \textsuperscript{6}University of Southern California \quad
  \textsuperscript{7}Washington University in St.\ Louis \\
  \textsuperscript{8}Shanghai Jiao Tong University \quad
  \textsuperscript{9}East China Normal University \quad
  \textsuperscript{10}Stanford University \\
  \textsuperscript{11}University of California, Berkeley \quad
  \textsuperscript{12}University of Michigan \quad
  \textsuperscript{13}Carnegie Mellon University \\[4pt]
}

\newcommand{\fix}{\marginpar{FIX}}
\newcommand{\new}{\marginpar{NEW}}

\begin{document}

\maketitle
\begin{abstract}
Perceptual constancy is the ability to maintain stable perceptions of objects despite changes in sensory input, such as variations in distance, angle, or lighting. This ability is crucial for visual understanding in a dynamic world. Here, we explored such ability in current Vision Language Models (VLMs). In this study, we evaluated 155 VLMs using 236 experiments across three domains: color, size, and shape constancy. The experiments included single-image and video adaptations of classic cognitive tasks, along with novel tasks in in-the-wild conditions. We found significant variability in VLM performance across these domains, with model performance in shape constancy clearly dissociated from that of color and size constancy.
\end{abstract}

\section{Introduction}
Perceptual constancy is the ability to perceive the properties of environmental objects as stable despite variations in external conditions. While this phenomenon exists across multiple sensory modalities, including auditory and tactile perception, it is particularly well-studied in vision. The human visual system demonstrates remarkable constancy, maintaining stable object perception despite changes in distance, viewing angle, lighting \citep{Epstein1977, Walsh1998}. This stability allows humans to perceive objects quickly and accurately, even in complex and dynamic scenes, ensuring seamless interaction with the surrounding world.

The study of perceptual constancy dates back to the 19th century, with early theories proposing that it emerges from the brain’s inference process—where stable perception is achieved by integrating empirical knowledge and environmental cues \citep{von1867handbuch}. Neuroscientific research has since demonstrated that visual information is processed hierarchically, from the retina to the visual cortex (e.g., V1, V4, IT cortex), to maintain perceptual stability across changing conditions \citep{dicarlo2012}. For example, neurons in the IT cortex consistently encode color regardless of illumination, suggesting that different levels of the visual system progressively compensate for environmental variations to preserve object properties \citep{tanaka1996}. This intricate mechanism enables humans to navigate dynamic environments, recognize objects across varying perspectives, and make accurate judgments in complex real-world scenarios. Similarly, instilling AI models with human-like perceptual constancy is crucial for developing robust vision-language models (VLMs) that can perform reliably under diverse and unpredictable conditions. Given the rapid progress of VLMs, perceptual constancy therefore appears to be a key benchmark for assessing their cognitive capabilities and identifying their limitations.

Perceptual constancy is essential for overcoming visual ambiguities in everyday life. Color constancy allows us to recognize objects consistently despite changes in lighting—a critical ability for autonomous systems in varying illumination conditions, such as self-driving cars interpreting traffic signals at different times of day \citep{article}. Size constancy ensures that objects appear the same size regardless of distance, facilitating spatial awareness and depth perception. This is particularly important for AI applications in robotics, where accurate size perception is necessary for grasping and object manipulation \citep{carlson2010}. Shape constancy enables stable object recognition across different viewpoints, supporting real-world tasks such as facial recognition, medical imaging, and augmented reality applications \citep{sternberg2006}. Without these constancy mechanisms, AI systems may struggle with perceptual inconsistencies, leading to unreliable in-the-wild performance that lacks adaptability and precision.

To systematically assess the perceptual constancy capabilities of VLMs, we leveraged the ConstancyBench from the \textbf{CoreCognition} benchmark \citep{li2024core}, formulating a dataset comprising 253 cognitive experiments. These experiments specifically examine the three primary dimensions of perceptual constancy: color, size, and shape. By testing 155 VLMs on these fundamental perceptual tasks, we aim to uncover the extent to which current models exhibit perceptual stability and where they fall short. Understanding these limitations is essential for advancing AI’s ability to process visual information in a way that aligns more closely with human perception, ultimately enabling more reliable and adaptable AI systems in real-world applications.

\begin{figure*}[t]
\centering
\includegraphics[width=1.0\textwidth]{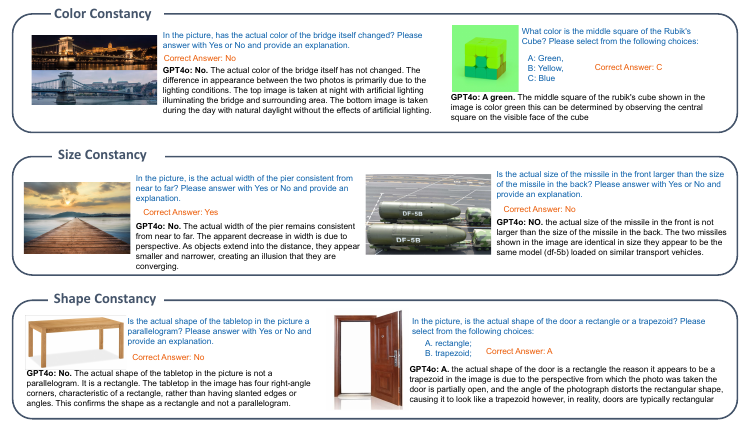}
\caption{\textbf{Sample Tasks from the Three Evaluation Dimensions of ConstancyBench.} Example model performance from GPT-4o is presented.}
\label{fig:fig1}
\end{figure*}

\section{Methods}

\subsection{Experiment Design}
Perceptual constancy refers to the human ability to perceive object properties consistently across varying environmental conditions, here focusing on three key domains: color, size, and shape constancy. These three aspects were selected to capture the fundamental principles of constancy in visual perception, enabling stable object recognition despite changes in lighting, distance, and viewing angle. Below, we provided explanations for each domain.

\subsubsection{Color Constancy}
Color constancy is an important feature of the human visual system. It allows us to perceive the color of objects consistently under different light conditions. A common example is a white wall appearing in different shades under different lighting conditions. However, the human visual system still perceives it as white rather than the wall's actual color has changed. This occurs because the visual system can separate an object's true color from the influence of lighting condition, thereby maintaining stable color perception \citep{Jameson1989}. Evaluating color constancy can reveal whether VLMs can truly understand an object's intrinsic color rather than merely relying on color patterns in the training data.

\subsubsection{Size Constancy}
Size constancy refers to the perception of an object’s size as stable, even when its retinal image changes due to variations in distance \citep{Sperandio2015}. For example, a distant car appears just as large as a nearby one, despite the difference in retinal projection. This stability is crucial for spatial awareness, depth perception, and navigation. Assessing this phenomenon in VLMs can determine whether they truly grasp the spatial properties of objects in the dynamic environment.

\subsubsection{Shape Constancy}
Shape constancy allows us to recognize objects as having the same shape, even when viewed from different angles \citep{Rock1973}. A round plate, for example, may project an elliptical image when seen obliquely, yet we still perceive it as circular. This perceptual stability relies on depth cues, prior experience, and contextual information. Shape constancy is fundamental to object recognition and spatial reasoning, allowing for accurate identification across perspectives. Research suggests humans achieve this by comparing novel views to stored shape representations \citep{tarr1995}. Evaluating shape constancy in VLMs can reveal whether they truly understand an object’s three-dimensional form or rely on fixed representations contingent to certain viewpoints.

\subsection{Examined Vision Language Models}
Recent advances in multi-modal learning have been driven by the unified modeling of visual and textual modalities using transformers \citep{li2019visualbert, xu2023bridgetower,tan2019lxmert, alayrac2022flamingo,radford2021learning}. With the rise of large language models (LLMs), state-of-the-art (SOTA) multi-modal LLMs (MLLMs) \citep{liu2024visual,li2023blip2} adopt open-source LLMs \citep{touvron2023llama, peng2023instruction,jiang2023mistral} and align visual features to the LLM embedding space \citep{li2023blip, fu2023mme, wu2024v, xu2024llava, shao2024visual, li2022more, li2025egoprivacy, brown2020language, achiam2023gpt, bai2023qwen, jaech2024openai, zhang2025unified, zhang2024pixels}. Progressively, MLLMs have demonstrated competitive performance in complex tasks involving high-level perception and reasoning \citep{li2024seed, team2023gemini, fu2023mme, openai2023gpt4}, such as spatial reasoning \citep{chen2024spatialvlm, cai2024spatialbot}, character recognition \citep{mori1999optical}, scene understanding \citep{cordts2016cityscapes, wang2023consistent, li2023diverse, chen2017deeplab}, action recognition \citep{jhuang2013towards, herath2017going} and prediction \citep{lan2014hierarchical, kong2022human}, reaching near-human performance.

We evaluated a total of 155 models for perceptual constancy analysis. Model performances are reported across three domains—color, size, and shape constancy. Each model was tested under a zero-shot setting, generating answers and textual explanations for each experimental prompt. Model size data and architecture type were recorded for correlation analysis.

\subsection{Data Sources}
The dataset includes six types of data sources: photographs (videos) taken, images (videos) from past classic cognitive psychology experiments, movies (animations), hand-drawn works, and AI-generated images (videos). The distribution is as follows:
\begin{enumerate}
\item There are 223 instances of photographs (videos);
\item There are 25 instances of images (videos) from classic cognitive psychology experiments;
\item There are 5 AI-generated images, involving standard 3D geometric shapes only.
\end{enumerate}
Some photographs and video materials have undergone post-processing for experimental purposes.

\section{Results}

\subsection{General Results}

Across 155 evaluated Vision-Language Models (VLMs), perceptual constancy performance varies substantially both across and within architectures. 
Accuracy spans a wide range—from below 0.20 in lightweight open-source models to above 0.90 in frontier systems such as GPT-4o and Gemini 1.5 Pro—highlighting a striking heterogeneity in perceptual stability. 
This variability underscores that perceptual constancy remains an emergent rather than universal capability: while certain large models exhibit localized robustness under specific transformations, no current architecture demonstrates consistent generalization across all constancy dimensions.

\begin{figure*}[t]
\centering
\includegraphics[width=0.5\textwidth]{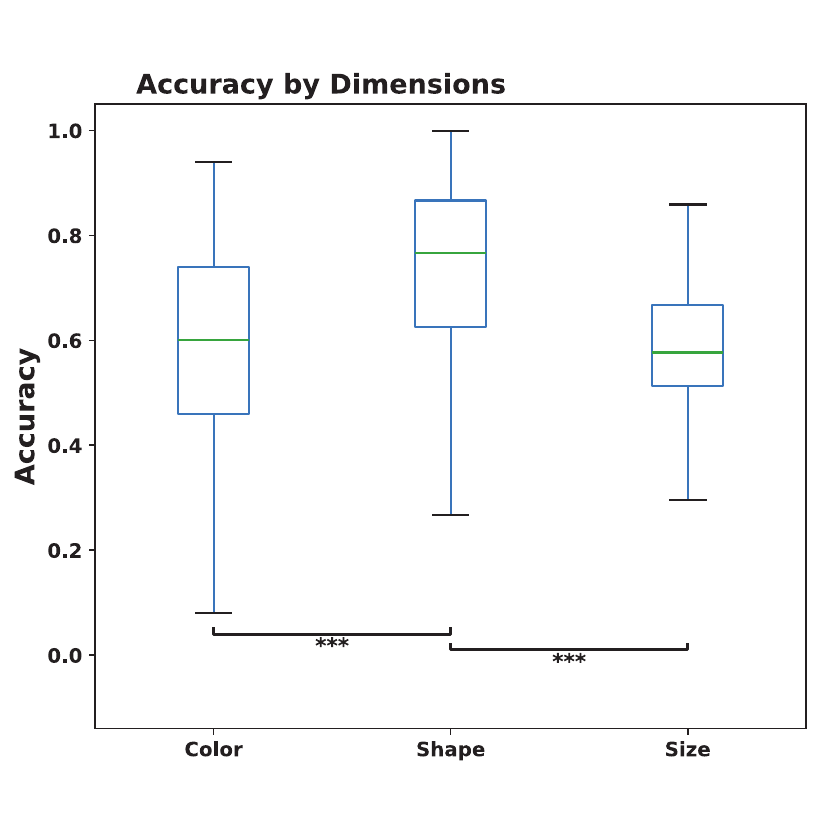}
\caption{
Bar plots show accuracy scores across vision--language models for color ($n=153$), shape ($n=152$), and size ($n=149$) constancy tasks after outlier removal 
(Color: $0.588 \pm 0.185$; Shape: $0.723 \pm 0.170$; Size: $0.584 \pm 0.123$). 
Horizontal bars with asterisks indicate statistical significance from post-hoc Tukey HSD tests 
(\textasteriskcentered~$p < 0.001$; ns = not significant, $p > 0.05$).
}

\label{fig:tukey_heatmap}
\end{figure*}

\subsection{Domain-wise Differences and Statistical Analysis}

Figure \ref{fig:tukey_heatmap} summarizes model performance across the three perceptual constancy domains.
VLMs demonstrate a clear performance hierarchy, achieving highest accuracy on shape constancy ($M=0.723$, $SD=0.170$), followed by color constancy ($M=0.588$, $SD=0.185$) and size constancy ($M=0.584$, $SD=0.123$).
Statistical analysis confirmed these domain-specific differences through a one-way ANOVA revealing a robust main effect of perceptual domain on accuracy ($F(2,451)=36.49$, $p=2.05\times10^{-15}$, $\eta^2=0.139$).
Post-hoc Tukey HSD comparisons demonstrated that shape constancy performance significantly exceeded both color ($p<0.001$, Cohen's $d=0.78$) and size constancy ($p<0.001$, Cohen's $d=0.89$), while color and size constancy showed statistically equivalent performance ($p=0.976$, Cohen's $d=0.02$).

These findings suggest shape constancy exploits simple geometric shortcuts, making it the most tractable domain for current VLMs. In contrast, color constancy requires high-dimensional photometric representations, while size constancy demands 3D world representation—explaining why shape dominates while color and size remain computationally challenging across models.

\subsection{Relationship Between Model Performance and Model Size}

\begin{figure*}[t]
\centering
\includegraphics[width=1.0\textwidth]{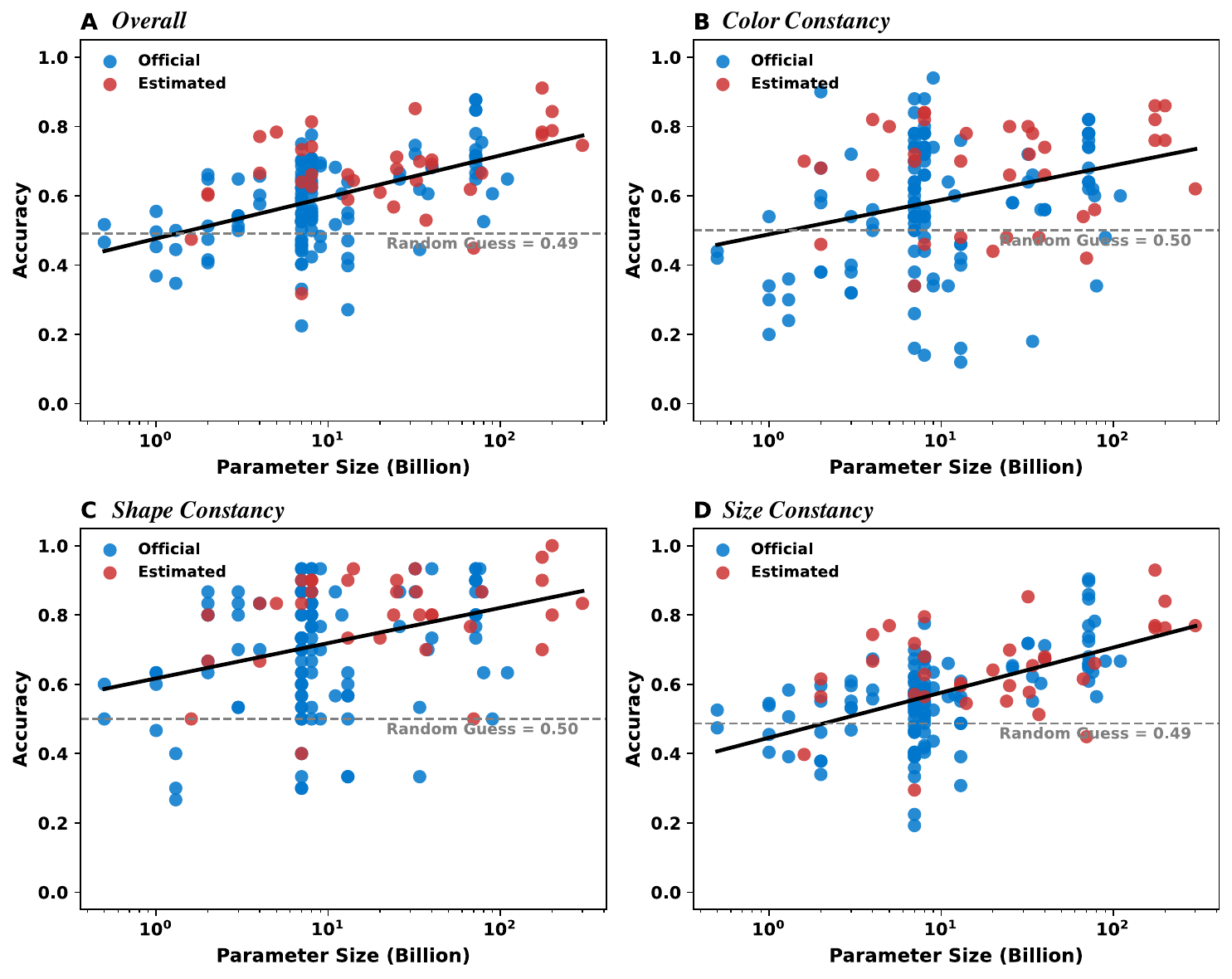}
\caption{\textbf{Relationship between model size and performance.} Larger models tend to perform better in perceptual constancy tasks.}
\label{fig:fig5}
\end{figure*}

To evaluate how architectural scale influences perceptual constancy, we examined the relationship between model performance and parameter count (Figure~\ref{fig:fig5}). 
Across 151 VLMs for which parameter estimates were available, regression analyses revealed a robust and consistent scaling effect. 
Overall accuracy increased significantly with model size (\(R^2 = 0.2804\), \(t = 7.62\), \(p = 2.74 \times 10^{-12}\)), following the log-linear relation \(y = 0.1200x + 0.4766\) where \(x = \log_{10}(\text{param})\).
Similar positive correlations were observed in all three perceptual constancy dimensions: 
color constancy (\(R^2 = 0.0973\), \(t = 4.01\), \(p = 9.66 \times 10^{-5}\), \(y = 0.0993x + 0.4888\)), 
shape constancy (\(R^2 = 0.1148\), \(t = 4.40\), \(p = 2.08 \times 10^{-5}\), \(y = 0.1017x + 0.6171\)), 
and size constancy (\(R^2 = 0.3080\), \(t = 8.14\), \(p = 1.42 \times 10^{-13}\), \(y = 0.1301x + 0.4457\)). 

These results collectively demonstrate that larger VLMs tend to achieve systematically higher perceptual stability across all constancy domains. 
The scaling relationships are not only statistically significant but also qualitatively consistent: as model capacity increases, representations become progressively more invariant to color, size, and shape perturbations. 
Among the three dimensions, the strongest scaling effect was observed in \textbf{size constancy}, indicating that geometric and spatial reasoning benefits most directly from expanded model capacity. 
In contrast, \textbf{color constancy} showed a weaker but still reliable dependence on size, suggesting that photometric invariance emerges more gradually with scale. 
Taken together, these findings support a unified scaling law of perceptual stability: 
\textit{the larger the model, the greater its resilience to perceptual variation}—a trend that spans all examined dimensions without evidence of domain-specific exceptions.

\subsection{Item Response Theory (IRT) Analysis}

To examine the latent structure underlying VLM performance across perceptual constancy tasks, we fitted a two-parameter logistic (2PL) Item Response Theory (IRT) model to all evaluated items. 
This framework characterizes each item by its discrimination parameter (\(a\)), which quantifies how sensitively the item differentiates between higher- and lower-ability models, and its difficulty parameter (\(b\)), which represents the ability level required to achieve a 50\% success probability. 

\begin{figure*}[t]
\centering
\includegraphics[width=1.0\textwidth]{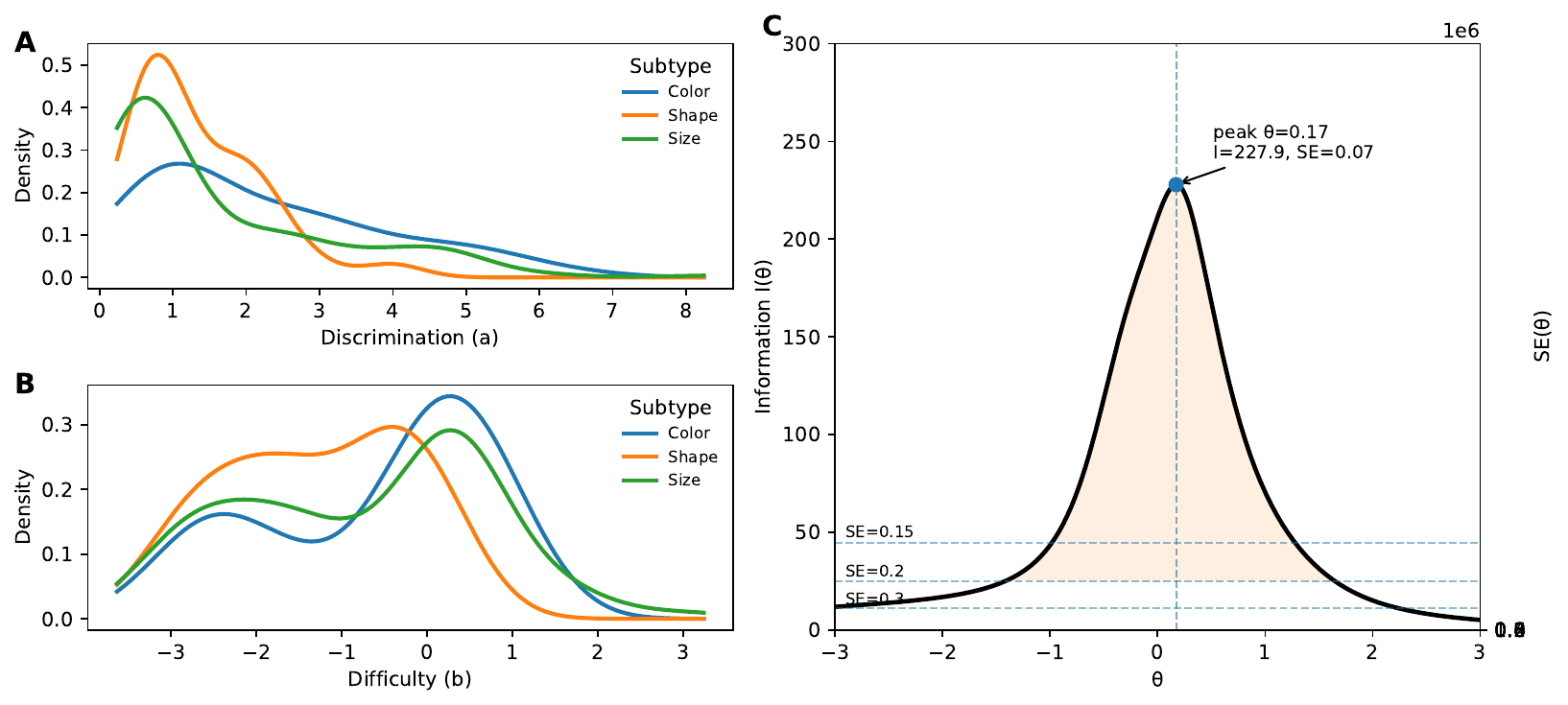}
\caption{\textbf{Item Response Theory (IRT) analysis of model-level performance.} 
Each point represents a single item parameterized by discrimination (\(a\)) and difficulty (\(b\)).
Shape-constancy items cluster in regions of lower difficulty and moderate discrimination, whereas color and size items show wider dispersion, reflecting greater variability in perceptual demands and model sensitivity.}
\label{fig:figIRT}
\end{figure*}

Across the three perceptual domains, clear structural distinctions emerged (Figure~\ref{fig:figIRT}A and Figure~\ref{fig:figIRT}B). 
\textbf{Color} items exhibited the highest mean discrimination (\(\bar{a}=2.25,\,SD=1.62\)) and moderate difficulty (\(\bar{b}=-0.58,\,SD=1.32\)), indicating that color-related tasks sharply separate stronger from weaker models, but require only moderate ability for success. 
\textbf{Shape} items showed lower discrimination (\(\bar{a}=1.33,\,SD=0.84\)) combined with the lowest mean difficulty (\(\bar{b}=-1.27,\,SD=1.08\)), reflecting tasks that most models could solve reliably and consistently—consistent with their high observed pass rate (\(0.76\)). 
In contrast, \textbf{Size} items demonstrated intermediate discrimination (\(\bar{a}=1.59,\,SD=1.56\)) and the broadest difficulty distribution (\(\bar{b}=-0.72,\,SD=1.47\)), suggesting greater heterogeneity in spatial reasoning demands and item complexity within this domain.

The global test information curve (Figure~\ref{fig:figIRT}C) peaks around \(\theta \approx 0\), indicating maximal measurement precision for models of average ability, with standard error (SE) below 0.20 across a wide ability range (\(-1.2 < \theta < +1.1\)). 
Together, these results reveal a well-calibrated hierarchy of perceptual challenges: 
shape constancy tasks anchor the lower end of the difficulty continuum and provide stable measurement of baseline competence, 
while color and size constancy tasks contribute discriminative power at moderate to high ability levels. 
This latent structure mirrors behavioral accuracy trends.

\section{Discussion}

This study introduces \textbf{ConstancyBench}, a large-scale benchmark for evaluating perceptual constancy in 155 Vision-Language Models (VLMs) across three fundamental domains—color, size, and shape constancy. 
Our results reveal that perceptual constancy is not a uniform emergent ability but a stratified one. 

\textbf{Hierarchical emergence of constancy.}  
The clear dissociation between shape versus color and size constancy supports the existence of a representational hierarchy in how VLMs internalize perceptual stability. 
Shape constancy—rooted in geometric invariance and low-level topology—appears to emerge from the structural regularities encoded within early convolutional or transformer layers. 
Such mechanisms correspond to what has been termed \textit{minimal constancy}: the capacity to preserve stable representations through accumulated perceptual regularities rather than explicit reasoning about context \citep{bradley2008constancy, buccella2021problem, buccella2022reconsidering}. 
By contrast, color and size constancy require broader contextual integration, including illumination estimation, depth reasoning, and perspective correction—abilities that depend more heavily on multimodal fusion and scene-level abstraction. 
This hierarchical division parallels the cognitive progression observed in human perception, where geometric stability precedes the development of higher-order contextual invariances.

\textbf{Scaling dynamics and emergent thresholds.}  
Our scaling analyses reveal that improvements in constancy performance are systematic yet nonlinear. 
Regression trends indicate that all three domains benefit from increased model capacity, but the relationship is strongest for size constancy, which shows a steep performance gain beyond approximately 10–13 billion parameters. 
This pattern echoes the “emergent threshold” phenomena widely observed in large language models, where complex behaviors—such as compositional reasoning or in-context adaptation—abruptly appear once sufficient representational depth is achieved. 
Analogously, perceptual constancy in VLMs may follow a \textit{phase-transition dynamic}, where representational stability reorganizes once the network’s capacity surpasses a critical complexity threshold. 
This finding bridges cognitive theories of perceptual maturation with the scaling laws of modern foundation models.

\textbf{Cognitive interpretation and representational implications.}  
From a cognitive perspective, the observed hierarchy mirrors developmental patterns in human perception: infants acquire shape constancy early through sensorimotor interaction, whereas color and size constancy emerge later as the visual system integrates cross-modal cues. 
VLMs seem to recapitulate this trajectory at the algorithmic level—progressing from local geometric consistency toward contextual and relational invariance. 
However, their mechanism of acquisition is fundamentally different: rather than inferential generalization from embodied experience, these models achieve constancy via statistical regularities embedded in massive multimodal datasets. 
This raises an important conceptual question: do VLMs exhibit \textit{functional constancy}—genuine invariance to physical transformations—or merely \textit{correlational constancy}, a byproduct of data coverage and alignment? 
Addressing this distinction will be key to evaluating whether these systems possess perceptual understanding or only statistical resemblance to it.

\textbf{Toward mechanistic understanding and future directions.}  
Mechanistically, probing how constancy-related invariances are encoded across visual and cross-attentional layers could reveal whether these patterns reflect bottom-up feature persistence or top-down integration. 
Architecturally, scaling alone may be insufficient: inductive biases such as explicit 3D priors, lighting normalization, or causal simulation modules may be required to support genuine invariance. 
Extending \textbf{ConstancyBench} to video, embodied, and generative contexts could further assess how temporal continuity and physical interaction contribute to perceptual stability. 
Such extensions would enable evaluation of \textit{causal perceptual reasoning}—the ability not only to maintain perceptual constancy but also to predict and explain variation under changing conditions. 
Ultimately, perceptual constancy offers a powerful diagnostic lens: it operationalizes the transition from mere perception to world modeling, providing a measurable bridge between cognitive theory and the representational dynamics of modern AI.

\section{Conclusion}

This work establishes \textbf{ConstancyBench} as the first systematic framework for probing perceptual constancy in large Vision-Language Models (VLMs). 
Through 236 controlled experiments spanning color, size, and shape constancy, we reveal that current VLMs exhibit a stratified rather than unified form of perceptual stability. 
Shape constancy emerges robustly even in smaller models, whereas color and size constancy depend strongly on scale and multimodal integration capacity. 
Our findings uncover a hierarchical pattern of perceptual emergence—one that mirrors developmental trajectories in human perception—and demonstrate that perceptual robustness follows a quantifiable scaling law across modern multimodal systems. 

Beyond benchmarking, this study suggests perceptual constancy as a cognitive lens for evaluating the \emph{world-modeling fidelity} of foundation models. 
Future work should examine how such invariances arise within model representations, whether through implicit statistical regularities or explicit geometric grounding, and extend these analyses to embodied and dynamic settings. 
Understanding and mechanistically modeling perceptual constancy will be central to bridging the gap between visual perception and true physical understanding in artificial intelligence.

\newpage

\bibliographystyle{plainnat}
\bibliography{iclr2025_conference}

\newpage 
\appendix

\section{Appendix}

\subsection{Model accuracy and IRT-derived ability parameters across perceptual dimensions.}
\label{app:irt_tables}
\begin{longtable}{lccccc}
\caption{}\\
\toprule
\textbf{Model} & \textbf{Accuracy} & \textbf{$\theta_{total}$} & \textbf{$\theta_{color}$} & \textbf{$\theta_{shape}$} & \textbf{$\theta_{size}$} \\
\midrule
\endfirsthead
\toprule
\textbf{Model} & \textbf{Accuracy} & \textbf{$\theta_{total}$} & \textbf{$\theta_{color}$} & \textbf{$\theta_{shape}$} & \textbf{$\theta_{size}$} \\
\midrule
\endhead

o1 & 0.911 & 1.707 & 0.820 & 0.967 & 0.929 \\
Qwen2.5-VL-72B-Instruct & 0.877 & 1.394 & 0.780 & 0.933 & 0.897 \\
Eagle-X5-13B-Chat & 0.640 & 1.366 & 0.950 & 0.900 & 0.683 \\
Qwen2.5-VL-72B-Instruct\_video & 0.877 & 1.361 & 0.780 & 0.900 & 0.904 \\
qwen-vl-max & 0.852 & 1.356 & 0.800 & 0.933 & 0.881 \\
gemini-1.5-pro & 0.788 & 1.307 & 0.860 & 0.800 & 0.763 \\
gemini-1.5-flash & 0.814 & 1.118 & 0.820 & 0.900 & 0.795 \\
Qwen2-VL-72B-Instruct & 0.847 & 1.083 & 0.760 & 0.933 & 0.859 \\
MiniCPM-Llama3-V-2\_5 & 0.742 & 1.013 & 0.840 & 0.900 & 0.679 \\
Eagle-X5-34B-Chat & 0.619 & 0.983 & 0.825 & 0.800 & 0.698 \\
gpt-4-turbo & 0.847 & 0.979 & 0.820 & 0.900 & 0.846 \\
gpt-4o & 0.843 & 0.902 & 0.760 & 1.000 & 0.840 \\
LLaVA-NeXT-Video-7B\_multi\_frame & 0.403 & 0.873 & 0.857 & 0.562 & 0.519 \\
Janus-Pro-7B & 0.593 & 0.858 & 0.975 & 0.850 & 0.604 \\
h2ovl-mississippi-1b & 0.716 & 0.844 & 0.820 & 0.900 & 0.647 \\
xgen-mm-phi3-dpo-r-v1.5 & 0.784 & 0.816 & 0.800 & 0.833 & 0.769 \\
gemini-1.5-flash-8b & 0.775 & 0.801 & 0.720 & 0.867 & 0.776 \\
Llama-3-LongVILA-8B-512Frames & 0.500 & 0.766 & 0.780 & 0.533 & 0.404 \\
Ovis1.5-Gemma2-9B & 0.686 & 0.756 & 0.940 & 0.767 & 0.590 \\
LLaVA-Video-7B-Qwen2\_multi\_frame & 0.750 & 0.753 & 0.880 & 0.933 & 0.673 \\
LLaVA-NeXT-Video-7B-DPO\_multi\_frame & 0.445 & 0.732 & 0.842 & 0.600 & 0.500 \\
Kosmos2 & 0.475 & 0.728 & 0.700 & 0.500 & 0.400 \\
Mantis-8B-Fuyu & 0.555 & 0.721 & 0.780 & 0.714 & 0.468 \\
InternVL-Chat-V1-2 & 0.703 & 0.720 & 0.740 & 0.800 & 0.673 \\
xgen-mm-phi3-interleave-r-v1.5 & 0.771 & 0.712 & 0.820 & 0.833 & 0.744 \\
VideoChat2\_stage3\_Mistral\_7B & 0.623 & 0.690 & 0.796 & 0.800 & 0.538 \\
Llama-3-LongVILA-8B-1024Frames & 0.564 & 0.684 & 0.880 & 0.800 & 0.417 \\
Ovis1.5-Llama3-8B & 0.674 & 0.667 & 0.780 & 0.933 & 0.590 \\
LLaVA-NeXT-Video-32B-Qwen\_multi\_frame & 0.746 & 0.653 & 0.720 & 0.933 & 0.718 \\
llava\_next\_mistral\_7b & 0.665 & 0.652 & 0.680 & 0.867 & 0.622 \\
LLaVA-Video-7B-Qwen2 & 0.703 & 0.610 & 0.840 & 0.900 & 0.622 \\
LLaVA-Video-72B-Qwen2\_multi\_frame & 0.780 & 0.606 & 0.740 & 0.900 & 0.769 \\
InternVL2\_5-78B & 0.754 & 0.581 & 0.600 & 0.867 & 0.782 \\
claude-3-7-sonnet-20250219 & 0.708 & 0.571 & 0.714 & 0.867 & 0.757 \\
Llama-3-LongVILA-8B-256Frames & 0.542 & 0.568 & 0.800 & 0.733 & 0.423 \\
InternVL2-8B-MPO-CoT & 0.708 & 0.562 & 0.740 & 0.833 & 0.673 \\
InternVL2-8B-MPO & 0.708 & 0.562 & 0.740 & 0.833 & 0.673 \\
llava-onevision-qwen2-72b-ov-chat-hf & 0.708 & 0.562 & 0.796 & 0.733 & 0.679 \\
llava-onevision-qwen2-72b-ov-hf & 0.691 & 0.561 & 0.740 & 0.767 & 0.673 \\
h2ovl-mississippi-2b & 0.513 & 0.559 & 0.600 & 0.633 & 0.462 \\
Ovis1.6-Gemma2-9B & 0.695 & 0.555 & 0.740 & 0.933 & 0.639 \\
mPLUG-Owl3 & 0.640 & 0.543 & 0.700 & 0.900 & 0.571 \\
claude-3-opus-20240229 & 0.775 & 0.542 & 0.860 & 0.700 & 0.763 \\
claude-3-5-sonnet-20240620 & 0.784 & 0.540 & 0.760 & 0.900 & 0.769 \\
llava-onevision-qwen2-7b-ov-hf & 0.686 & 0.537 & 0.633 & 0.733 & 0.699 \\
deepseek-vl2-small & 0.733 & 0.532 & 0.720 & 0.833 & 0.718 \\
yi-vision-v2 & 0.699 & 0.528 & 0.780 & 0.800 & 0.654 \\
Mantis-8B-Idefics2 & 0.674 & 0.516 & 0.740 & 0.933 & 0.603 \\
XinYuan-VL-2B-Instruct & 0.648 & 0.511 & 0.680 & 0.867 & 0.596 \\
VideoLLaMA2-72B & 0.716 & 0.489 & 0.640 & 0.800 & 0.724 \\
InternVL-Chat-V1-2-Plus & 0.691 & 0.453 & 0.660 & 0.800 & 0.679 \\
LLaVA-Video-72B-Qwen2 & 0.733 & 0.451 & 0.640 & 0.867 & 0.737 \\
video\_chat\_7b & 0.458 & 0.450 & 0.587 & 0.655 & 0.473 \\
grok-2-vision-1212 & 0.746 & 0.427 & 0.620 & 0.833 & 0.769 \\
llava-llama-3-8b & 0.606 & 0.411 & 0.660 & 0.800 & 0.551 \\
Qwen2-VL-7B-Instruct & 0.699 & 0.398 & 0.700 & 0.933 & 0.654 \\
MiniCPM-V-2\_6 & 0.661 & 0.395 & 0.840 & 0.867 & 0.564 \\
MMAlaya2 & 0.712 & 0.377 & 0.660 & 0.867 & 0.699 \\
llava-onevision-qwen2-7b-ov-chat-hf & 0.606 & 0.375 & 0.633 & 0.733 & 0.588 \\
LLaVA-NeXT-Video-32B-Qwen & 0.720 & 0.368 & 0.640 & 0.867 & 0.718 \\
Ovis1.6-Llama3.2-3B & 0.648 & 0.343 & 0.720 & 0.800 & 0.596 \\
Llama-3.2V-11B-cot & 0.682 & 0.336 & 0.640 & 0.867 & 0.660 \\
Aria & 0.678 & 0.329 & 0.800 & 0.900 & 0.596 \\
llava-onevision-qwen2-72b-si-hf & 0.665 & 0.326 & 0.620 & 0.800 & 0.662 \\
Phi-4-multimodal-instruct & 0.644 & 0.300 & 0.780 & 0.933 & 0.545 \\
Qwen2-VL-2B-Instruct & 0.661 & 0.298 & 0.900 & 0.833 & 0.551 \\
llava\_next\_72b & 0.648 & 0.275 & 0.680 & 0.800 & 0.609 \\
Mantis-8B-siglip-llama3 & 0.623 & 0.262 & 0.660 & 0.833 & 0.574 \\
Eagle-X4-13B-Plus & 0.551 & 0.246 & 0.575 & 0.950 & 0.633 \\
llava-onevision-qwen2-7b-si-hf & 0.576 & 0.225 & 0.653 & 0.700 & 0.539 \\
InternVL2\_5-26B & 0.665 & 0.220 & 0.580 & 0.867 & 0.654 \\
Llama-3-VILA1.5-8B-Fix & 0.665 & 0.192 & 0.560 & 0.767 & 0.679 \\
pllava-7b & 0.551 & 0.187 & 0.571 & 0.750 & 0.540 \\
hunyuan-vision & 0.661 & 0.183 & 0.700 & 0.900 & 0.603 \\
Phi-3.5-Vision & 0.665 & 0.166 & 0.660 & 0.667 & 0.667 \\
Idefics3-8B-Llama3 & 0.636 & 0.153 & 0.720 & 0.767 & 0.583 \\
VILA1.5-40B & 0.682 & 0.152 & 0.560 & 0.733 & 0.712 \\
InternVL-Chat-V1-5 & 0.665 & 0.147 & 0.560 & 0.867 & 0.660 \\
VideoLLaMA2-7B & 0.470 & 0.145 & 0.440 & 0.567 & 0.462 \\
VideoChat2\_HD\_stage4\_Mistral\_7B\_hf & 0.475 & 0.144 & 0.580 & 0.667 & 0.412 \\
Video-LLaVA-7B & 0.525 & 0.125 & 0.700 & 0.533 & 0.468 \\
InternVL2\_5-38B & 0.606 & 0.124 & 0.560 & 0.700 & 0.603 \\
llava\_next\_110b & 0.648 & 0.123 & 0.600 & 0.633 & 0.667 \\
qwen-vl-plus-2025-01-25 & 0.644 & 0.121 & 0.720 & 0.867 & 0.596 \\
Qwen2.5-VL-7B-Instruct & 0.653 & 0.116 & 0.760 & 0.867 & 0.577 \\
VideoLLaMA2-7B-16F & 0.593 & 0.109 & 0.580 & 0.733 & 0.571 \\
Qwen2.5-VL-7B-Instruct\_video & 0.644 & 0.098 & 0.740 & 0.867 & 0.571 \\
internlm-xcomposer2d5-7b & 0.331 & 0.089 & 0.641 & 0.818 & 0.473 \\
InternVL2-40B & 0.682 & 0.080 & 0.560 & 0.933 & 0.673 \\
InternVL2-8B & 0.644 & 0.080 & 0.540 & 0.800 & 0.647 \\
deepseek-vl-7b-chat & 0.525 & 0.078 & 0.463 & 0.900 & 0.626 \\
video\_chat\_13b & 0.419 & 0.066 & 0.457 & 0.607 & 0.466 \\
InternVL2-76B & 0.669 & 0.060 & 0.620 & 0.933 & 0.635 \\
InternVL2\_5-1B & 0.555 & 0.040 & 0.540 & 0.633 & 0.545 \\
emu2-chat & 0.530 & 0.029 & 0.480 & 0.700 & 0.513 \\
LLaVA-Video-7B-Qwen2-Video-Only\_multi\_frame & 0.504 & 0.028 & 0.520 & 0.700 & 0.462 \\
Pixtral-12B-2409 & 0.606 & 0.027 & 0.600 & 0.800 & 0.571 \\
llava\_next\_llama3 & 0.449 & 0.025 & 0.525 & 0.750 & 0.504 \\
InternVL2-26B & 0.648 & 0.015 & 0.580 & 0.767 & 0.647 \\
VILA1.5-3b & 0.513 & 0.012 & 0.540 & 0.700 & 0.468 \\
SmolVLM & 0.602 & 0.007 & 0.680 & 0.667 & 0.564 \\
LLaVA-NeXT-Video-7B & 0.441 & 0.000 & 0.804 & 0.556 & 0.382 \\
gpt-4o-mini & 0.627 & -0.035 & 0.460 & 0.900 & 0.628 \\
InternVL2-4B & 0.657 & -0.043 & 0.500 & 0.833 & 0.673 \\
InternVL2\_5-8B & 0.551 & -0.053 & 0.520 & 0.767 & 0.519 \\
Mini-InternVL-Chat-4B-V1-5 & 0.602 & -0.065 & 0.520 & 0.833 & 0.583 \\
Janus-1.3B & 0.445 & -0.071 & 0.450 & 0.400 & 0.568 \\
llava\_next\_interleave\_7b\_dpo & 0.525 & -0.096 & 0.540 & 0.733 & 0.481 \\
deepseek-vl2 & 0.619 & -0.102 & 0.540 & 0.767 & 0.615 \\
deepseek-vl2-tiny & 0.606 & -0.107 & 0.460 & 0.800 & 0.615 \\
internlm-xcomposer2-7b & 0.623 & -0.109 & 0.580 & 0.828 & 0.614 \\
deepseek-vl-1.3b-chat & 0.500 & -0.109 & 0.375 & 0.600 & 0.655 \\
Llama-3.2-90B-Vision-Instruct & 0.606 & -0.115 & 0.480 & 0.500 & 0.667 \\
InternVL2-2B & 0.475 & -0.124 & 0.580 & 0.800 & 0.378 \\
Llama-3-VILA1.5-8B & 0.534 & -0.126 & 0.440 & 0.633 & 0.545 \\
InternVL2\_5-4B & 0.576 & -0.133 & 0.560 & 0.724 & 0.558 \\
Mantis-8B-clip-llama3 & 0.530 & -0.142 & 0.480 & 0.700 & 0.513 \\
LLaVA-NeXT-Video-7B-DPO & 0.453 & -0.142 & 0.646 & 0.556 & 0.418 \\
llava-onevision-qwen2-0.5b-ov-hf & 0.517 & -0.142 & 0.440 & 0.600 & 0.526 \\
InternVL-Chat-V1-1 & 0.589 & -0.149 & 0.480 & 0.733 & 0.596 \\
LLaVA-Video-7B-Qwen2-Video-Only & 0.462 & -0.152 & 0.540 & 0.633 & 0.404 \\
claude-3-sonnet-20240229 & 0.568 & -0.161 & 0.480 & 0.800 & 0.555 \\
llava-onevision-qwen2-0.5b-si-hf & 0.466 & -0.171 & 0.420 & 0.500 & 0.484 \\
Qwen2.5-VL-3B-Instruct & 0.542 & -0.205 & 0.380 & 0.867 & 0.532 \\
pllava-13b & 0.470 & -0.205 & 0.444 & 0.556 & 0.524 \\
Llama-3-LongVILA-8B-128Frames & 0.542 & -0.216 & 0.540 & 0.833 & 0.487 \\
Qwen2.5-VL-3B-Instruct\_video & 0.542 & -0.220 & 0.400 & 0.833 & 0.532 \\
claude-3-haiku-20240307 & 0.610 & -0.223 & 0.440 & 0.733 & 0.641 \\
llava\_next\_interleave\_7b & 0.530 & -0.238 & 0.540 & 0.533 & 0.526 \\
Chat-UniVi-13B & 0.271 & -0.269 & 0.750 & 0.625 & 0.522 \\
VideoLLaMA2.1-7B-16F & 0.585 & -0.296 & 0.500 & 0.600 & 0.609 \\
qwen\_chat & 0.508 & -0.385 & 0.380 & 0.700 & 0.513 \\
VILA1.5-13b & 0.534 & -0.449 & 0.460 & 0.567 & 0.551 \\
InternVL2\_5-2B & 0.415 & -0.455 & 0.380 & 0.667 & 0.378 \\
Qwen2.5-Omni-7B & 0.547 & -0.468 & 0.340 & 0.733 & 0.577 \\
Eagle-X5-34B-Plus & 0.445 & -0.523 & 0.225 & 0.500 & 0.619 \\
Mini-InternVL-Chat-2B-V1-5 & 0.407 & -0.551 & 0.380 & 0.800 & 0.342 \\
Vintern-3B-beta & 0.538 & -0.587 & 0.320 & 0.533 & 0.609 \\
Eagle-X4-8B-Plus & 0.424 & -0.601 & 0.184 & 0.750 & 0.561 \\
idefics\_9b\_instruct & 0.453 & -0.677 & 0.383 & 0.724 & 0.500 \\
idefics\_80b\_instruct & 0.525 & -0.686 & 0.347 & 0.633 & 0.564 \\
JanusFlow-1.3B & 0.347 & -0.692 & 0.300 & 0.450 & 0.455 \\
VILA1.5-3b-s2 & 0.500 & -0.702 & 0.320 & 0.533 & 0.551 \\
Llama-3.2-11B-Vision-Instruct & 0.517 & -0.707 & 0.340 & 0.567 & 0.575 \\
OpenFlamingo-9B-vitl-mpt7b & 0.483 & -0.725 & 0.340 & 0.500 & 0.526 \\
Janus-Pro-1B & 0.369 & -0.734 & 0.244 & 0.700 & 0.450 \\
InternVL2-1B & 0.453 & -0.787 & 0.340 & 0.633 & 0.455 \\
Vintern-1B-v2 & 0.496 & -0.787 & 0.306 & 0.600 & 0.538 \\
Eagle-X5-7B & 0.403 & -0.862 & 0.200 & 0.450 & 0.561 \\
Eagle-X5-13B & 0.398 & -0.872 & 0.200 & 0.500 & 0.547 \\
Valley2-7b & 0.051 & -0.882 & 0.286 & 0.500 & 0.267 \\
video\_chatgpt-7B & 0.225 & -0.926 & 0.271 & 0.357 & 0.210 \\
Chat-UniVi & 0.318 & -0.950 & 0.340 & 0.414 & 0.297 \\
qwen\_base & 0.140 & -1.023 & 0.400 & 0.417 & 0.558 \\
\bottomrule
\end{longtable}

\newpage

\subsection{Item-level discrimination and difficulty parameters estimated from the 2PL IRT model.}

\begin{longtable}{rccccc}
\caption{ \(a\) = discrimination, \(b\) = difficulty, passrate = proportion correct.}\\
\toprule
\textbf{Item ID} & \textbf{Subtype} & \textbf{$a$} & \textbf{$b$} & \textbf{Passrate} \\
\midrule
\endfirsthead
\toprule
\textbf{Item ID} & \textbf{Subtype} & \textbf{$a$} & \textbf{$b$} & \textbf{Passrate} \\
\midrule
\endhead

 400488 & Size   &  8.241 &   0.204 &  0.503 \\
 400462 & Color  &  6.436 &  -0.327 &  0.820 \\
 400487 & Size   &  6.178 &   0.238 &  0.480 \\
 400486 & Size   &  5.615 &   0.172 &  0.520 \\
 400515 & Color  &  5.434 &   0.056 &  0.597 \\
 400500 & Color  &  5.430 &  -0.293 &  0.793 \\
 400502 & Color  &  5.338 &  -0.260 &  0.779 \\
 400444 & Size   &  5.184 &   0.261 &  0.464 \\
 400449 & Color  &  4.969 &  -0.263 &  0.773 \\
 400440 & Size   &  4.904 &   0.189 &  0.507 \\
 400475 & Size   &  4.855 &  -0.376 &  0.818 \\
 400458 & Size   &  4.848 &   0.230 &  0.487 \\
 400441 & Size   &  4.846 &   0.335 &  0.423 \\
   1732 & Size   &  4.793 &  -0.055 &  0.656 \\
 400459 & Size   &  4.753 &  -0.445 &  0.840 \\
 400484 & Size   &  4.676 &   0.124 &  0.550 \\
 400467 & Color  &  4.646 &  -0.307 &  0.788 \\
 400507 & Size   &  4.598 &   0.373 &  0.388 \\
 400504 & Color  &  4.552 &   0.041 &  0.603 \\
 400442 & Size   &  4.539 &   0.480 &  0.336 \\
 400482 & Size   &  4.485 &   0.224 &  0.490 \\
 400513 & Size   &  4.417 &   0.362 &  0.405 \\
 400453 & Color  &  4.263 &  -0.055 &  0.651 \\
 400499 & Size   &  4.234 &   0.098 &  0.563 \\
 400472 & Size   &  4.219 &   0.064 &  0.587 \\
 400517 & Color  &  4.104 &  -0.295 &  0.772 \\
 400470 & Size   &  4.023 &   0.242 &  0.477 \\
 400456 & Size   &  3.995 &  -0.496 &  0.838 \\
   1740 & Shape  &  3.977 &   0.056 &  0.583 \\
 400509 & Size   &  3.873 &   0.224 &  0.490 \\
   1731 & Size   &  3.790 &   0.290 &  0.446 \\
 400501 & Color  &  3.690 &   0.440 &  0.369 \\
   1745 & Size   &  3.672 &   0.259 &  0.467 \\
 400446 & Size   &  3.667 &   0.253 &  0.477 \\
 400468 & Color  &  3.551 &   0.351 &  0.423 \\
   1762 & Size   &  3.396 &   0.108 &  0.550 \\
 400523 & Color  &  3.385 &   0.395 &  0.397 \\
 400438 & Size   &  3.380 &   0.468 &  0.360 \\
   1861 & Size   &  3.376 &  -0.072 &  0.649 \\
   1744 & Color  &  3.213 &   0.036 &  0.593 \\
 400471 & Size   &  3.212 &   0.907 &  0.174 \\
   1765 & Size   &  3.142 &   0.448 &  0.368 \\
   1730 & Color  &  3.074 &   1.063 &  0.132 \\
 400437 & Color  &  3.034 &   0.523 &  0.340 \\
   1736 & Size   &  2.875 &   0.478 &  0.365 \\
 400520 & Color  &  2.857 &  -0.026 &  0.616 \\
   1776 & Size   &  2.830 &   1.140 &  0.121 \\
   1826 & Shape  &  2.825 &  -0.724 &  0.874 \\
   1733 & Color  &  2.806 &   0.283 &  0.460 \\
 400495 & Size   &  2.754 &   0.324 &  0.443 \\
   1803 & Size   &  2.662 &   0.562 &  0.329 \\
   1798 & Size   &  2.641 &   0.323 &  0.440 \\
 400524 & Color  &  2.622 &   0.342 &  0.433 \\
   1742 & Size   &  2.616 &  -0.419 &  0.772 \\
   1812 & Size   &  2.587 &   0.278 &  0.461 \\
   1852 & Size   &  2.560 &   0.516 &  0.351 \\
 400477 & Size   &  2.551 &   0.408 &  0.404 \\
   1882 & Color  &  2.511 &  -0.629 &  0.829 \\
   1814 & Size   &  2.505 &   0.687 &  0.283 \\
 400466 & Color  &  2.482 &   0.591 &  0.327 \\
   1821 & Size   &  2.466 &  -0.130 &  0.651 \\
   1817 & Shape  &  2.447 &  -0.177 &  0.667 \\
   1802 & Size   &  2.407 &   0.334 &  0.439 \\
 400522 & Color  &  2.373 &   0.497 &  0.368 \\
   1858 & Size   &  2.339 &   0.776 &  0.259 \\
   1804 & Color  &  2.330 &   0.452 &  0.388 \\
 400490 & Size   &  2.282 &   0.225 &  0.487 \\
   1875 & Size   &  2.281 &   0.260 &  0.473 \\
   1763 & Size   &  2.275 &   0.405 &  0.409 \\
   1743 & Color  &  2.268 &   1.018 &  0.185 \\
   1707 & Shape  &  2.177 &  -0.730 &  0.834 \\
   1800 & Size   &  2.175 &   0.457 &  0.385 \\
   1250 & Shape  &  2.157 &  -0.295 &  0.713 \\
   1783 & Size   &  2.104 &   0.376 &  0.430 \\
   1241 & Shape  &  2.087 &  -0.339 &  0.726 \\
   1853 & Shape  &  2.040 &  -0.497 &  0.763 \\
   1756 & Color  &  2.034 &   0.120 &  0.533 \\
   1784 & Size   &  1.946 &  -0.146 &  0.632 \\
   1874 & Shape  &  1.908 &  -1.248 &  0.914 \\
   1210 & Size   &  1.861 &   0.352 &  0.444 \\
   1747 & Size   &  1.856 &   0.714 &  0.309 \\
   1248 & Size   &  1.838 &   0.690 &  0.321 \\
   1813 & Shape  &  1.835 &   0.252 &  0.480 \\
   1836 & Shape  &  1.786 &  -0.009 &  0.576 \\
   1223 & Size   &  1.740 &   0.024 &  0.574 \\
   1240 & Size   &  1.681 &  -0.143 &  0.630 \\
   1805 & Size   &  1.630 &   0.711 &  0.325 \\
 400521 & Color  &  1.562 &   0.993 &  0.253 \\
 400489 & Size   &  1.553 &   0.901 &  0.278 \\
 400525 & Color  &  1.547 &   0.999 &  0.253 \\
   1247 & Shape  &  1.512 &   0.106 &  0.537 \\
1001208 & Size   &  1.458 &  -1.811 &  0.941 \\
 400519 & Color  &  1.416 &   0.807 &  0.318 \\
   1737 & Color  &  1.394 &   0.893 &  0.296 \\
   1221 & Shape  &  1.378 &  -0.457 &  0.699 \\
   1901 & Color  &  1.362 &  -1.141 &  0.841 \\
   1816 & Shape  &  1.357 &  -0.431 &  0.682 \\
   1208 & Size   &  1.356 &  -1.368 &  0.881 \\
   1242 & Color  &  1.345 &   0.272 &  0.482 \\
   1237 & Color  &  1.332 &   0.495 &  0.415 \\
1001723 & Size   &  1.310 &   0.479 &  0.424 \\
   1725 & Size   &  1.216 &   1.547 &  0.181 \\
   1888 & Color  &  1.200 &   0.146 &  0.513 \\
1001724 & Color  &  1.194 &  -2.524 &  0.959 \\
   1724 & Color  &  1.190 &  -3.082 &  0.980 \\
   1773 & Size   &  1.174 &   2.061 &  0.114 \\
2001723 & Size   &  1.164 &   0.646 &  0.388 \\
   1894 & Size   &  1.145 &   0.656 &  0.382 \\
 400436 & Size   &  1.124 &   1.252 &  0.250 \\
   1883 & Size   &  1.088 &   1.091 &  0.289 \\
   1791 & Size   &  1.081 &   0.515 &  0.423 \\
   1746 & Color  &  1.061 &   0.945 &  0.322 \\
   1212 & Shape  &  1.052 &  -2.385 &  0.934 \\
   1224 & Color  &  1.039 &  -1.650 &  0.867 \\
1001725 & Size   &  1.009 &   1.826 &  0.174 \\
1001704 & Shape  &  0.987 &  -0.860 &  0.733 \\
   1217 & Shape  &  0.967 &  -1.744 &  0.861 \\
   1222 & Color  &  0.954 &  -2.588 &  0.933 \\
   1751 & Size   &  0.923 &  -2.097 &  0.887 \\
   1220 & Size   &  0.915 &  -1.449 &  0.815 \\
   1779 & Size   &  0.914 &  -2.191 &  0.893 \\
   1244 & Size   &  0.878 &  -2.728 &  0.927 \\
   1831 & Shape  &  0.871 &   0.333 &  0.473 \\
3001723 & Size   &  0.870 &   0.126 &  0.517 \\
   1245 & Size   &  0.859 &   0.956 &  0.350 \\
 400445 & Size   &  0.852 &  -2.913 &  0.932 \\
   1704 & Shape  &  0.845 &  -1.695 &  0.827 \\
   1225 & Size   &  0.844 &  -2.779 &  0.925 \\
   1249 & Size   &  0.841 &  -2.301 &  0.890 \\
   1723 & Size   &  0.839 &   0.470 &  0.449 \\
   1230 & Size   &  0.839 &  -2.388 &  0.896 \\
   1239 & Color  &  0.823 &  -2.250 &  0.881 \\
   1228 & Color  &  0.820 &  -1.720 &  0.825 \\
   1209 & Size   &  0.820 &  -2.427 &  0.895 \\
   1215 & Shape  &  0.814 &  -2.115 &  0.867 \\
1001773 & Size   &  0.812 &   2.859 &  0.107 \\
   1699 & Color  &  0.807 &  -2.586 &  0.901 \\
   1227 & Size   &  0.789 &  -1.469 &  0.787 \\
 400478 & Size   &  0.785 &  -2.164 &  0.861 \\
   1232 & Shape  &  0.779 &  -3.015 &  0.926 \\
   1698 & Color  &  0.779 &  -2.635 &  0.899 \\
   1785 & Shape  &  0.766 &  -2.463 &  0.882 \\
   1702 & Size   &  0.757 &  -3.188 &  0.928 \\
   1235 & Size   &  0.750 &  -1.174 &  0.735 \\
   1891 & Color  &  0.742 &   0.322 &  0.477 \\
   1238 & Shape  &  0.740 &  -2.622 &  0.890 \\
   1809 & Size   &  0.736 &  -2.294 &  0.860 \\
   1739 & Size   &  0.732 &  -2.738 &  0.894 \\
   1728 & Size   &  0.722 &  -1.973 &  0.824 \\
   1769 & Size   &  0.704 &  -1.618 &  0.778 \\
   1706 & Shape  &  0.703 &  -2.218 &  0.844 \\
   1863 & Shape  &  0.702 &  -1.711 &  0.789 \\
   1734 & Size   &  0.699 &  -2.748 &  0.886 \\
 400455 & Size   &  0.698 &  -1.764 &  0.793 \\
   1219 & Shape  &  0.684 &  -1.674 &  0.784 \\
   1703 & Size   &  0.682 &  -2.900 &  0.893 \\
 400457 & Size   &  0.682 &  -3.629 &  0.934 \\
   1788 & Size   &  0.668 &  -2.468 &  0.854 \\
   1903 & Size   &  0.667 &  -3.082 &  0.900 \\
   1758 & Size   &  0.667 &  -3.099 &  0.901 \\
   1793 & Size   &  0.667 &  -1.854 &  0.795 \\
   1226 & Size   &  0.664 &  -2.006 &  0.812 \\
 400448 & Size   &  0.661 &  -2.561 &  0.860 \\
   1246 & Color  &  0.659 &  -2.559 &  0.861 \\
   1771 & Size   &  0.656 &  -0.462 &  0.605 \\
 400480 & Size   &  0.651 &  -2.878 &  0.882 \\
   1700 & Color  &  0.642 &  -3.123 &  0.895 \\
   1229 & Color  &  0.642 &  -1.718 &  0.776 \\
   1218 & Color  &  0.640 &  -2.403 &  0.843 \\
2001773 & Size   &  0.634 &   3.240 &  0.127 \\
   1701 & Color  &  0.624 &  -2.788 &  0.867 \\
 400479 & Shape  &  0.620 &  -3.110 &  0.887 \\
   1708 & Size   &  0.615 &  -1.528 &  0.742 \\
2001725 & Size   &  0.615 &   1.499 &  0.313 \\
   1887 & Size   &  0.604 &  -3.012 &  0.875 \\
   1810 & Size   &  0.594 &  -1.291 &  0.707 \\
   1849 & Size   &  0.593 &  -2.856 &  0.860 \\
   1774 & Size   &  0.593 &  -3.036 &  0.873 \\
   1782 & Size   &  0.578 &  -1.556 &  0.733 \\
   1766 & Size   &  0.563 &  -1.345 &  0.705 \\
   1873 & Size   &  0.553 &  -0.680 &  0.619 \\
   1792 & Size   &  0.551 &  -2.343 &  0.803 \\
   1829 & Shape  &  0.538 &  -2.936 &  0.847 \\
   1868 & Size   &  0.537 &  -2.531 &  0.813 \\
   1764 & Size   &  0.531 &  -0.399 &  0.578 \\
   1906 & Size   &  0.525 &  -2.247 &  0.783 \\
   1825 & Shape  &  0.522 &  -2.507 &  0.807 \\
   1881 & Color  &  0.513 &  -2.324 &  0.787 \\
 400483 & Size   &  0.506 &  -1.149 &  0.664 \\
 400511 & Size   &  0.506 &  -2.526 &  0.800 \\
   1711 & Size   &  0.495 &  -1.912 &  0.740 \\
   1815 & Size   &  0.489 &  -1.442 &  0.691 \\
   1854 & Size   &  0.483 &  -3.171 &  0.840 \\
   1234 & Color  &  0.483 &  -1.318 &  0.679 \\
   1786 & Size   &  0.472 &  -1.392 &  0.680 \\
   1869 & Size   &  0.469 &  -2.487 &  0.781 \\
   1847 & Size   &  0.465 &  -2.165 &  0.753 \\
   1741 & Size   &  0.464 &  -1.165 &  0.653 \\
   1778 & Size   &  0.455 &  -2.041 &  0.737 \\
   1845 & Size   &  0.454 &  -1.819 &  0.716 \\
   1860 & Shape  &  0.444 &  -1.384 &  0.671 \\
   1781 & Size   &  0.440 &  -0.895 &  0.618 \\
   1801 & Size   &  0.438 &  -0.880 &  0.616 \\
   1738 & Size   &  0.437 &   1.258 &  0.384 \\
   1859 & Shape  &  0.433 &  -1.459 &  0.673 \\
   1844 & Size   &  0.432 &  -2.180 &  0.740 \\
   1880 & Size   &  0.425 &  -1.788 &  0.702 \\
   1885 & Size   &  0.422 &  -1.469 &  0.671 \\
   1895 & Size   &  0.420 &  -1.209 &  0.645 \\
   1770 & Size   &  0.417 &  -0.522 &  0.576 \\
   1797 & Size   &  0.410 &  -0.126 &  0.533 \\
   1871 & Size   &  0.410 &  -1.349 &  0.656 \\
   1710 & Size   &  0.409 &  -1.685 &  0.687 \\
   1897 & Size   &  0.402 &   1.797 &  0.342 \\
   1808 & Size   &  0.396 &  -0.699 &  0.589 \\
   1777 & Size   &  0.389 &  -1.002 &  0.617 \\
   1796 & Size   &  0.387 &   1.847 &  0.342 \\
 400473 & Size   &  0.385 &   0.570 &  0.463 \\
   1211 & Size   &  0.379 &  -2.272 &  0.726 \\
   1787 & Size   &  0.376 &   0.166 &  0.503 \\
   1780 & Size   &  0.372 &  -0.227 &  0.539 \\
   1748 & Size   &  0.366 &  -2.779 &  0.755 \\
   1902 & Size   &  0.365 &  -0.394 &  0.555 \\
   1856 & Size   &  0.365 &  -0.529 &  0.567 \\
   1790 & Size   &  0.365 &  -0.813 &  0.592 \\
   1729 & Size   &  0.363 &   1.273 &  0.401 \\
 400493 & Size   &  0.362 &  -2.697 &  0.747 \\
   1806 & Size   &  0.357 &   0.573 &  0.466 \\
   1840 & Size   &  0.351 &   0.001 &  0.517 \\
   1705 & Size   &  0.319 &  -1.680 &  0.651 \\
 400496 & Size   &  0.300 &  -2.999 &  0.733 \\
   1752 & Size   &  0.300 &  -3.221 &  0.747 \\
 400460 & Size   &  0.297 &  -2.894 &  0.725 \\
 400461 & Size   &  0.293 &  -3.380 &  0.752 \\
 400447 & Color  &  0.248 &  -2.001 &  0.642 \\
 400498 & Size   &  0.239 &  -1.672 &  0.617 \\
\bottomrule
\end{longtable}

\newpage

\subsection{Additional Examples}

\begin{figure*}[h]
\centering
\includegraphics[width=1.0\textwidth]{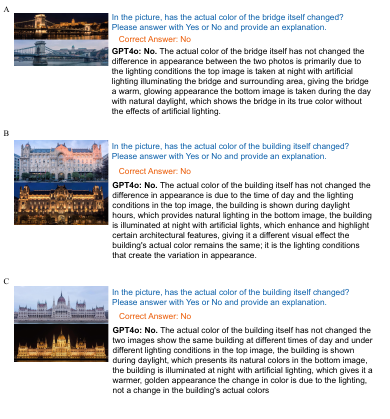}
\caption{\textbf{Additional Sample Tasks: Color Constancy}.}
\label{fig:fig31}
\end{figure*}

\begin{figure*}[h]
\centering
\includegraphics[width=1.0\textwidth]{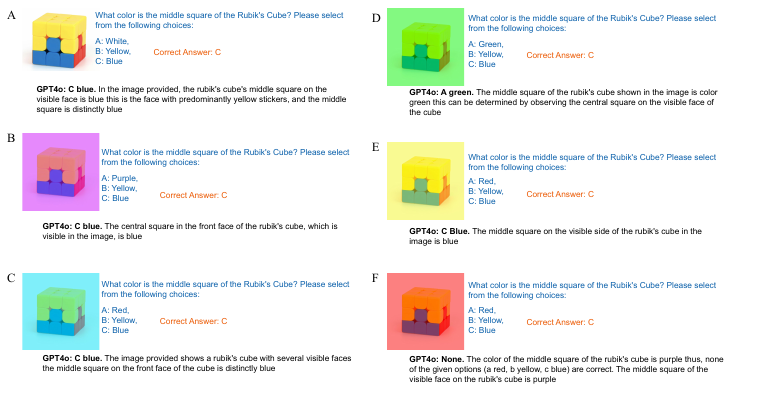}
\caption{\textbf{Additional Sample Tasks: Color Constancy}.}
\label{fig:fig32}
\end{figure*}

\begin{figure*}[h]
\centering
\includegraphics[width=1.0\textwidth]{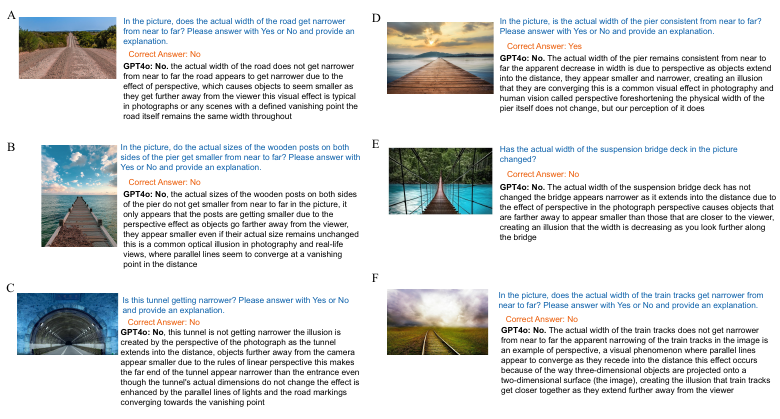}
\caption{\textbf{Additional Sample Tasks: Size Constancy}.}
\label{fig:fig33}
\end{figure*}

\begin{figure*}[h]
\centering
\includegraphics[width=1.0\textwidth]{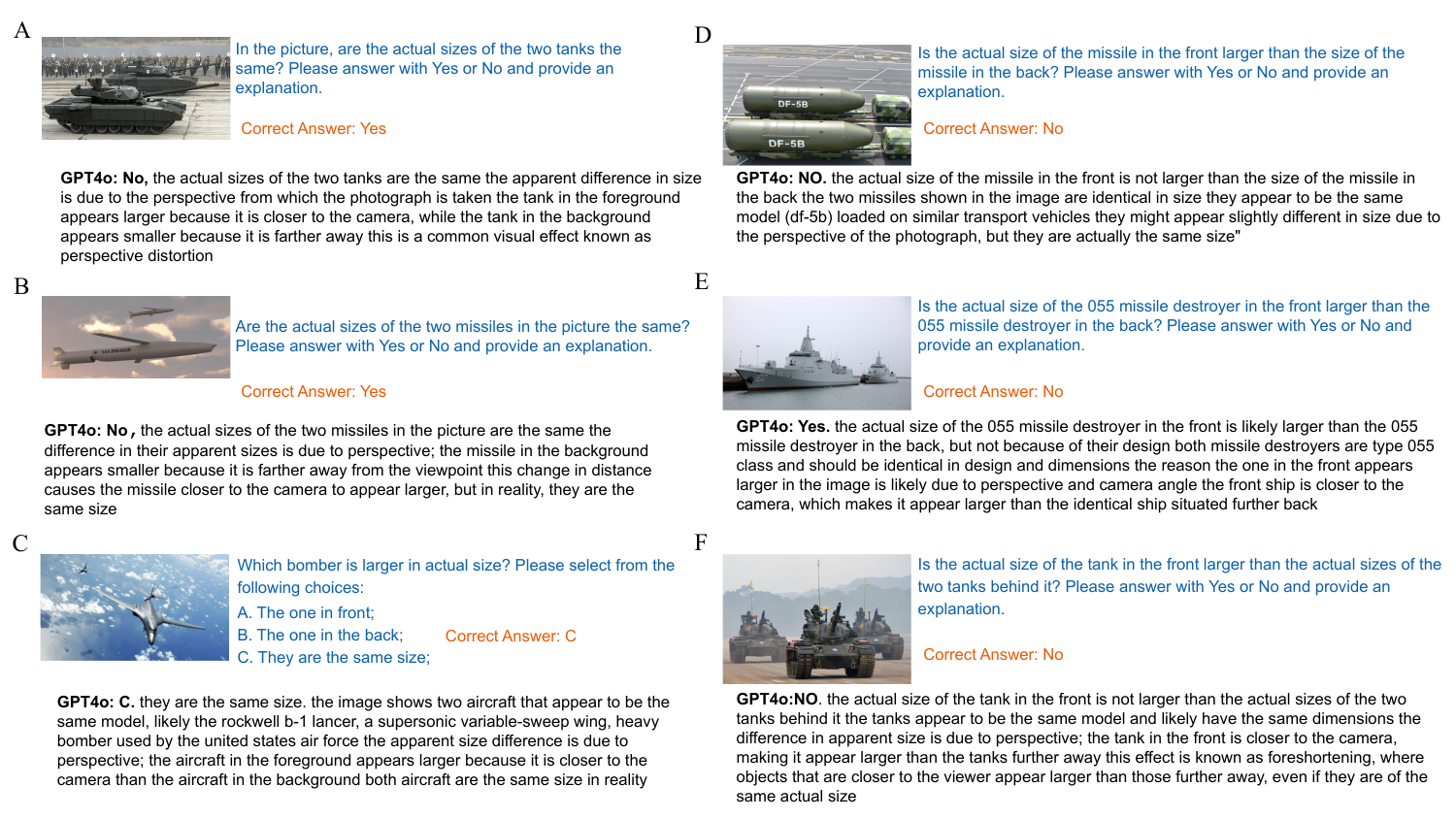}
\caption{\textbf{Additional Sample Tasks: Size Constancy}.}
\label{fig:fig34}
\end{figure*}

\begin{figure*}[h]
\centering
\includegraphics[width=1.0\textwidth]{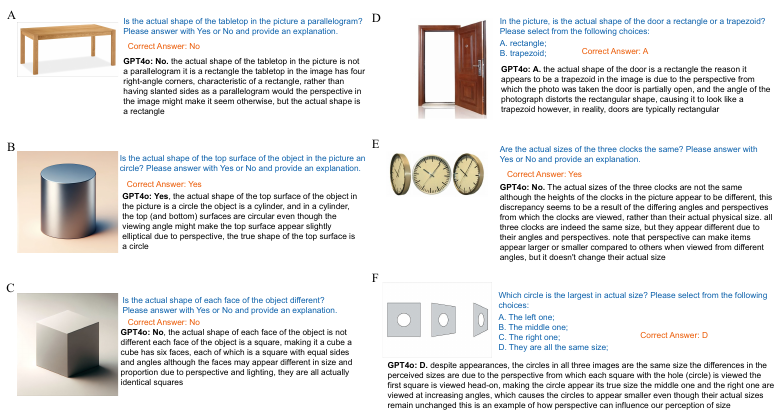}
\caption{\textbf{Additional Sample Tasks: Shape Constancy}.}
\label{fig:fig35}
\end{figure*}

\end{document}